\documentclass[letterpaper, 10 pt, conference]{ieeeconf}
\usepackage{cite}
\usepackage{amsmath,amssymb,amsfonts}
\usepackage{algorithmic}
\usepackage{graphicx}
\usepackage{textcomp}
\usepackage{xcolor}
\usepackage{subcaption}
\usepackage{hyperref}
\usepackage{fancyhdr}
\IEEEoverridecommandlockouts                              

\usepackage{multirow}

\newcommand{\etal}{\emph{et al.}}

\title{\LARGE \bf VaPr: Variable-Precision Tensors to Accelerate Robot Motion Planning
}
\author{
Yu-Shun Hsiao$^\dagger$, Siva Kumar Sastry Hari$^\ddagger$, Balakumar Sundaralingam$^\ddagger$, Jason Yik$^\dagger$, Thierry Tambe$^\dagger$,
\\ 
Charbel Sakr$^\ddagger$, Stephen W. Keckler$^\ddagger$, Vijay Janapa Reddi$^\dagger$  \\
$^\ddagger$ NVIDIA Corporation, $^\dagger$ Harvard University
    }

\begin{document}
\maketitle
\thispagestyle{empty}
\pagestyle{empty}


\begin{abstract}
High-dimensional motion generation requires numerical precision for smooth, collision-free solutions. Typically, double-precision or single-precision floating-point (FP) formats are utilized. Using these for big tensors imposes a strain on the memory bandwidth provided by the devices and alters the memory footprint, hence limiting their applicability to low-power edge devices needed for mobile robots. The uniform application of reduced precision can be advantageous but severely degrades solutions. Using decreased precision data types for important tensors, we propose to accelerate motion generation by removing memory bottlenecks. We propose variable-precision (VaPr) search optimization to determine the appropriate precision for large tensors from a vast search space of approximately 4 million unique combinations for FP data types across the tensors.
To obtain the efficiency gains, we exploit existing platform support for an out-of-the-box GPU speedup and evaluate prospective precision converter units for GPU types that are not currently supported.
Our experimental results on 800 planning problems for the Franka Panda robot on the MotionBenchmaker dataset across 8 environments show that a 4-bit FP format is sufficient for the largest set of tensors in the motion generation stack.
With the software-only solution, VaPr achieves 6.3\% and 6.3\% speedups on average for a significant portion of motion generation over the SOTA solution (CuRobo) on Jetson Orin and RTX2080 Ti GPU, respectively, and 9.9\%, 17.7\% speedups with the FP converter. 

\end{abstract}

\section{Introduction}
\label{intro}
Robot manipulators are increasingly used in the industry to work alongside humans, assisting humans in repetitive, and labor-intensive tasks.
These tasks often have slight variations between runs, thereby requiring online motion planning to complete tasks successfully. The state-of-the-art (SOTA) CPU-based motion planning methods (Tesseract~\cite{tesseract} and OMPL~\cite{sucan2012open}) take significant time to plan motions, ranging from seconds to minutes, depending on the environment. For example, Tesseract takes 3 seconds on average to generate motion in the 8 standard environments from MotionBenchMaker~\cite{chamzas2021motionbenchmaker}, with the 98 percentile planning time being 16 seconds. 

Recently, CuRobo ($0.3.1$ version) accelerated motion generation by 100$\times$ on NVIDIA RTX 3090 GPU and 50$\times$ on Jetson AGX Orin~\cite{curobo}. CuRobo achieved this speedup by exploiting parallelism, optimizing multiple trajectories simultaneously with different initialization on the GPU. CuRobo lowered the motion generation time to 25ms (or 40 Hz on the median on RTX 3090), bringing online motion planning closer to real-world deployment. However, when running CuRobo on a low-power embedded system, the planning time increases to 64 ms on median (on a NVIDIA Jetson AGX Orin). In addition, there will be other tasks running simultaneously on the embedded system, such as perception and control. Furthermore, faster than real-time motion generation can benefit emerging applications that simulate robots in virtual worlds (or digital twins). 


%

In this study, we investigate the performance characteristics of the CuRobo motion generation pipeline. Large tensors are proportional to batch size and they can be larger than the L2 cache size of SOTA systems, such as 4 MB for Jetson AGX Orin (and 4MB on RTX 2080). The costly data transfers to and from off-chip memory create performance bottlenecks. The memory footprint limits the scalability of batch  applications (e.g., faster online data production). Normally, double precision (FP64) is used in robot manipulators to attain a high success rate but at the expense of incurring a significant latency~\cite{sucan2012open}. Ichnowski~\etal \cite{ichnowski2019motion} showed that using FP32 can reduce the planning time. Our experiments indicate that adopting half-precision~(FP16) throughout the whole motion planning pipeline would render it incapable of generating suitable trajectories. Nonetheless, if we use variable-precision (VaPr) as we propose to optimize data types for large tensors of the critical kernels, we can reduce the memory footprint and eliminate memory bottlenecks without compromising the motion generation pipeline.

To reduce data transfers for large intermediate CUDA tensors, we propose VaPr searching for a feasible combination of reduced precision for these tensors. The search is challenging due to the enormous search space. Even after we narrow down to five tensors based on our performance characterization (Sec.~\ref{sec:characterization}) and limit the number of possible reduced-precision FP data formats to 21 choices (Sec.~\ref{subsec:search_space}), the search space still has 4,084,101 possibilities. As a result, we propose a novel VaPr technique to reduce the search space by up to 7.35$\times$ and sample the search space efficiently with an evolutionary algorithm, NSGA-II~\cite{deb2002fast}. After search space reduction, our results show that as small as 500 samples can provide a feasible result that, on average, shrinks the tensors by at least three times, reducing the total bitwidths of five FP32 tensors from 160 bits down to 43 bits or less. 


Despite the utilization of VaPr in the search for reduced precision within the CuRobo motion generation pipeline, it can also be applied to other pipelines with CUDA kernels. Developers can employ the VaPr search technique to minimize memory movements associated with larger intermediate tensors on the GPU. Moreover, the VaPr search technique is independent of the underlying hardware and only requires a single execution for a new environment. Additionally, this technique enables parallel sample evaluation and can be completed within a few minutes when utilizing industry-scale computing resources.

To realize the performance benefit with reduced precision using VaPr, this paper presents an out-of-box software-only solution and a hardware solution. 
We leverage the existing type-conversion support between FP32 and FP16 in CUDA for the software-only solution. 
To quantize further below FP16, this paper proposes the use of fast FP converter hardware units. 
Our experimental results show that VaPr-CuRobo accelerates the CuRobo motion generation pipeline by 6.3\%, 6.3\% on average with existing CUDA intrinsics support on Jetson Orin and RTX 2080 Ti GPU, respectively, and 9.9\%, 17.7\% with our proposed hardware support.



In summary, the contributions of this work are as follows:
\begin{itemize}
    \item The first characterization study of an end-to-end accelerated motion generation pipeline where we present the runtime breakdown for CuRobo motion generation. 
    \item VaPr-CuRobo to search for the optimal reduced-precision FP data types for the large tensors of the key kernels out of a huge search space.
    \item 6.3\% faster trajectory optimization and Inverse Kinematics (IK), commonly used portions of the motion generation pipeline, with an out-of-box software solution.
    \item Proposal to further speed up IK and trajectory optimization by 22.3\% and 13.0\%, respectively, with new FP conversion units.
\end{itemize}
\section{Background}
\label{sec:background}
\begin{figure}[t!]
        \centering\includegraphics[width=0.85\columnwidth]{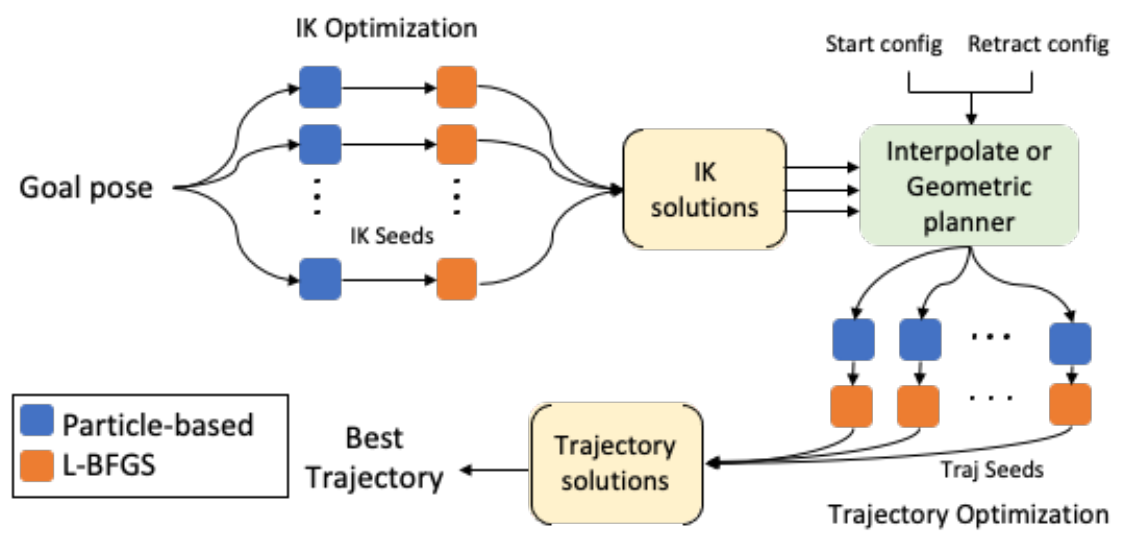}
        \vspace{-0.1in}
        \caption{CuRobo motion generation pipeline for manipulators.}
        \vspace{-0.2in}
        \label{fig:curobo}
\end{figure}

\subsection{CuRobo Motion Planning Pipeline} 

CuRobo is the SOTA GPU-accelerated motion generation pipeline. It produces superior paths (shorter in C-space) much faster than any of the prior CPU-based methods~\cite{kuffner2000rrt,strub2020adaptively,tesseract}. Specifically, CuRobo is about 100x and 50x faster on NVIDIA RTX 3090 and Jetson Orin AGX, respectively, compared to Tesseract~\cite{tesseract}. 
%
The CuRobo pipeline includes inverse kinematics (IK) optimization (IKO), interpolation-based and geometric planner (GP), and trajectory optimization (TO), as shown in Fig.~\ref{fig:curobo}. For the first step, IK optimization turns the goal pose of the robot's end effector $X_g$ into multiple optimized terminal joint configurations $\ddot{\theta}_t$ with random IK seeds in a batch. For each IK solution, the interpolation-based planner checks if steering from start to goal configuration directly or through a predefined retract configuration is possible. If not, the GP samples collision-free configurations from an informed search region within the straight line distance between start and goal, similar to BIT*~\cite{gammell2015batch}. In this paper, we always employ the geometric planner to better seed the trajectory optimization (eventually, a DNN trained for a specific environment can also be used to seed TO). Finally, each of the seed trajectories goes through trajectory optimization to make it smooth and optimized with respect to predefined cost functions. A single run of the motion generation pipeline is defined as one attempt. If there are no feasible trajectories for the current attempt, the whole motion generation pipeline will make another attempt with new sets of random seeds. A challenging task in a complex environment typically requires multiple attempts.

\subsection{CuRobo Kernels}

CuRobo provides CUDA kernels for robot forward and backward kinematics, robot-environment and robot-self distance queries, and the L-BFGS step direction computation. L-BFGS solver is a gradient-based optimization that computes the step direction towards a local optima of a cost function. Robot forward kinematics turns joint configuration into the Cartesian coordinate of the end effector. IK takes the Cartesian coordinate of the end effector as input and output a robot's joint configuration. Robot-environment and robot-self distance queries are utilized in the cost function to avoid collisions with itself or the environment.

\section{Characterization}
\label{sec:characterization}

\subsection{Experimental Setup}
Edge devices with power and form factor constraints are one of the target systems. 
We use a Jetson AGX Orin device and configure it to operate at 15W and 60W (referred to as MAXN) power budgets. In addition, we use a RTX 2080 Ti GPU for large-batched applications (e.g., fast online data generation). Table~\ref{tab:jetson} shows the devices we use in this paper along with their configurations, e.g., number of TPCs (one TPC has 2 streaming multi-processors or SMs) and clock frequencies for the GPU, DRAM, and CPU.

\renewcommand{\arraystretch}{1.2}
\begin{table}
\caption{GPUs and their power modes used for evaluation.}
\vspace{-5pt}
\resizebox{\columnwidth}{!}{
\begin{tabular}{|c|c|c|c|c|c| } 
\hline
\multirow{2}{*}{Device} & \multirow{2}{*}{Mode} & \multirow{2}{*}{\#TPCs} & \multicolumn{3}{|c|}{Frequency (MHz)} \\
\cline{4-6}
 & & & GPU & DRAM & CPU \\
\hline
\hline
\multirow{2}{9em}{Jetson Orin (Ampere GPU, Carmel CPU)} 
 & 15W & 3 & 421 & 2,133 & 1,114\\ \cline{2-6}
& MAXN & 8 & 1,301 & 3,199 & 2,202 \\ \hline

\multirow{2}{9em}{RTX 2080 Ti (Turing GPU, i9-9940X CPU)} 
 & \multirow{2}{*}{TDP 300W} & \multirow{2}{*}{34} & \multirow{2}{*}{2,190} & \multirow{2}{*}{Uncontrolled} & \multirow{2}{*}{Max 3,300} \\ 
 & & & & & \\\hline
\end{tabular}
}
\label{tab:jetson}
\vspace{-0.2in}
\end{table}

\subsection{Motion Generation Pipeline Characterization}
\begin{figure*}[t!]
  \subfloat[Orin 15W.]{
	\begin{minipage}[b][][b]{
	   0.32\linewidth}
	   \includegraphics[width=\textwidth]{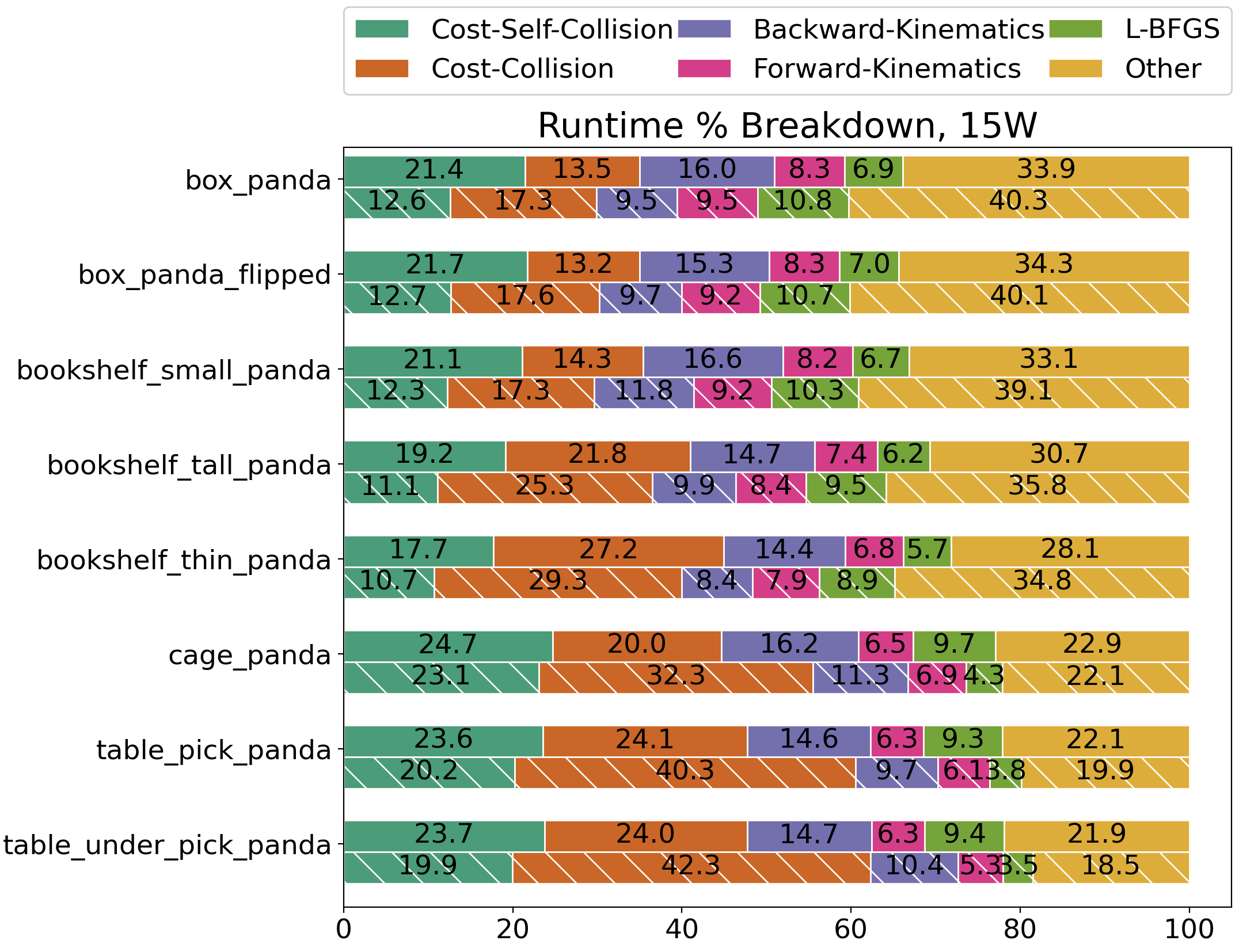}
	   \label{fig:15W_breakdown}
	   \vspace{-0.2in}
	\end{minipage}}
	  \subfloat[Orin MAXN.]{
	\begin{minipage}[b][][b]{
	   0.32\linewidth}
	   \includegraphics[width=\textwidth]{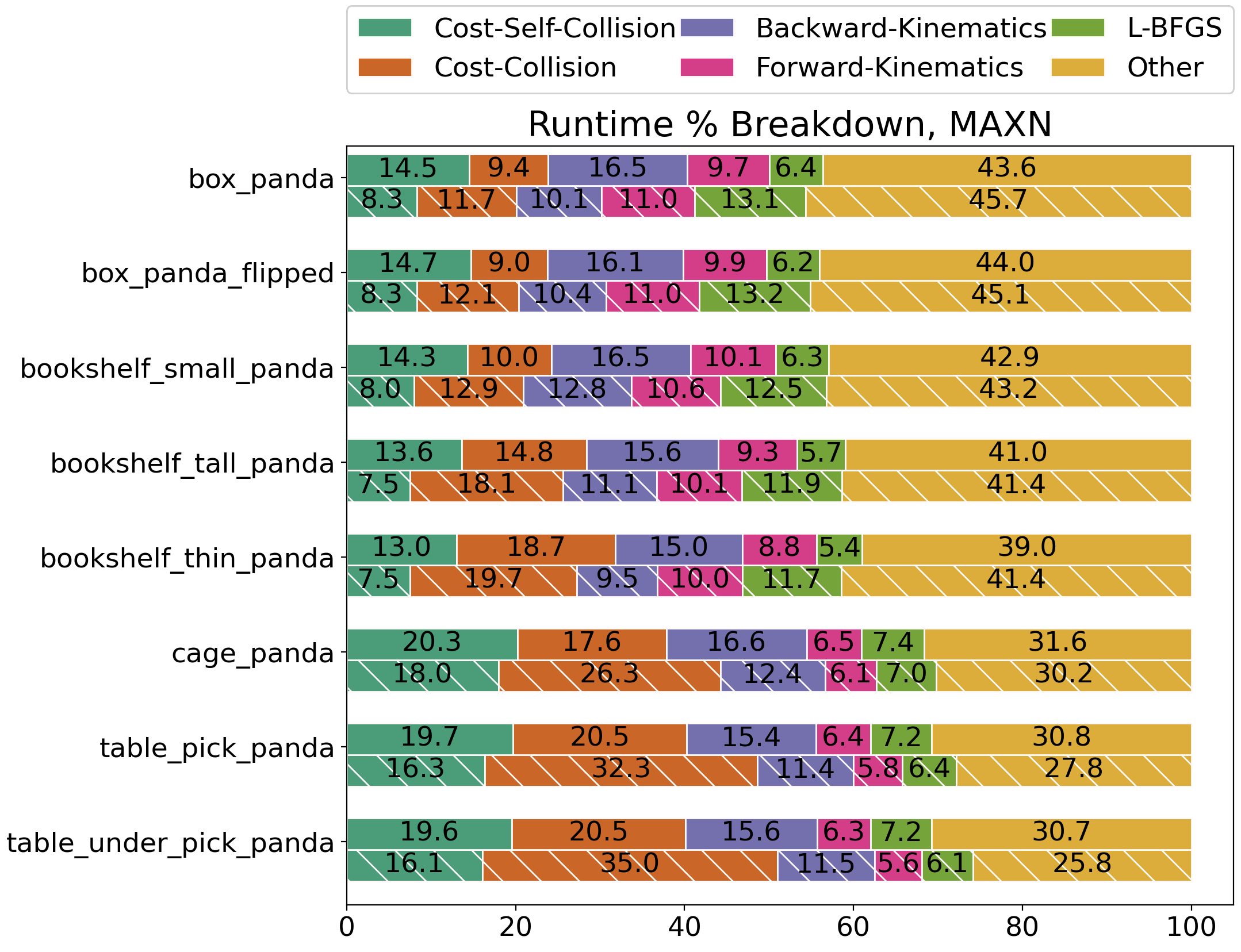}
	   \label{fig:MAXN_breakdown}
	   \vspace{-0.2in}
	\end{minipage}}
		  \subfloat[2080 Ti.]{
	\begin{minipage}[b][][b]{
	   0.32\linewidth}
	   \includegraphics[width=\textwidth]{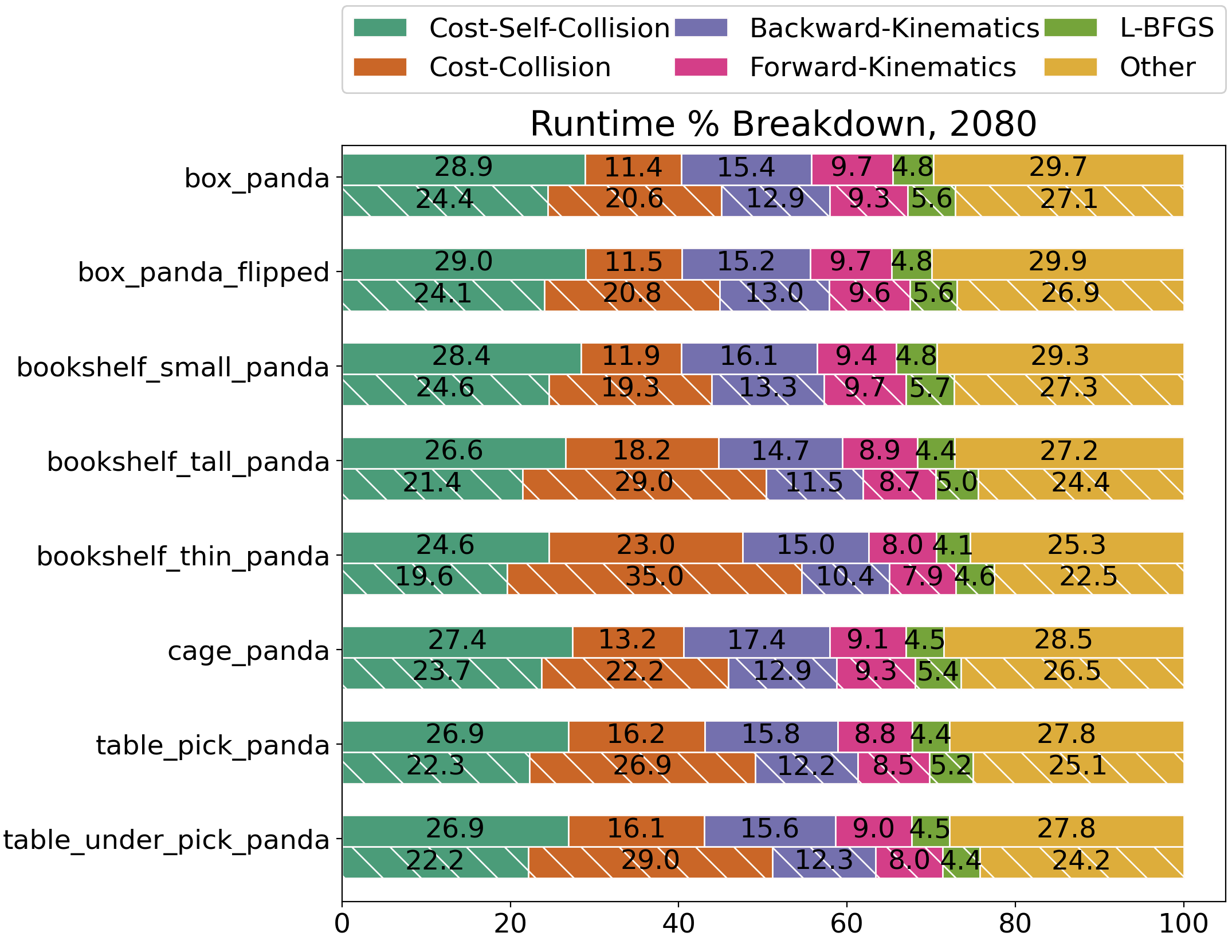}
	   \label{fig:2080_breakdown}
	   \vspace{-0.2in}
	\end{minipage}}
	\vspace{-0.05in}
\caption{GPU runtime breakdown of inverse kinematics optimization (IKO, solid bars) and trajectory optimization (TO, hashed bars on Orin and 2080Ti GPUs for different environments.}
\label{fig:breakdown}
\vspace{-15pt}
\end{figure*}


As described in Sec.~\ref{sec:background}, IKO, GP, and TO are the three main phases of motion generation. 
CuRobo shows that GP is not always required for a smooth collision-free plan generation; it is invoked sometimes, so the average runtime contribution is small. Moreover, GP can potentially be replaced by an environment-specific DNN trained to generate motion fast to seed trajectory optimization. So we focus on the IKO and TO phases in this paper and employ GP to seed TO in all the problems. The core components (e.g., kinematics and collision checking kernels) from IKO and TO are also used in GP. Hence, the optimizations presented in this paper should also benefit GP.


Each IKO and TO iteration performs the following high-level computations. (1) Given N, step scales of step direction (see (7)). (2) Compute kinematics. (3) Compute cost functions which include pose, collision, self-collision, and bound position. (4) Aggregating the costs. (5) Compute backward for the above steps in the reverse sequence. (6) Use line search to pick one from N. (7) Lastly, compute step direction (L-BFGS) and buffer updates. 


In Figure~\ref{fig:breakdown}, we show the GPU runtime breakdowns for various compute kernels aggregated across all iterations of IKO and TO for different environments at 15W and MAXN power modes on Orin and 2080 Ti GPU. The \textit{cage\_panda}, \textit{table\_pick\_panda}, and \textit{table\_under\_pick\_panda} environments represent challenging environments and use 400 IKO and 20 TO seeds configurations, while the other environments use 100 IKO and 2TO seeds configurations. We use 2000 IKO and 100 TO seeds for all environments on 2080 Ti to simulate large batch settings at data centers. The CUDA kernels include forward and backward kinematics, cost of self-collision and collision, and L-BFGS optimizer.

The takeaways from Figure~\ref{fig:breakdown} are the following.
(1) \textbf{The depicted kernels account for most of the runtime for both IKO and TO.}
Across all tested environments and system configurations, the depicted kernels' computing collision cost, self-collision cost, and forward and backward kinematics account for an average 68.49\% of total GPU runtime for IKO and 68.53\% for TO. Within the same environment, this sum is comparable between IKO and TO, at most different by 6\%. 
%
(2) \textbf{CUDA kernels take more percentage of the total runtime with a larger number of IKO and TO seeds on Orin.}
For the upper five environments with small seed settings, the depicted CUDA kernels account for 54.9\% to 71.9\% of the total GPU runtime on Orin. The percentage goes to at least 77.1\% for Orin 15W mode and  68.4\% for Orin MAXN mode for the lower three challenging environments with large seed settings. Since the batch size for the kernels scales with the number of optimization seeds, the CUDA kernels take more time than the other auto-generated kernels for small fragmented computations. Similarly, for 2080Ti, the combined CUDA kernels runtime concentrates toward the custom kernels and ranges from 70.3\% to 75.8\%.
(3) \textbf{Self-collision and collision checking runtimes vary across seed configurations.} For challenging environments that require more seeds, the combined runtime of self-collision and collision checking takes up to 47.7\% and 62.2\% of total runtime for IKO and TO, respectively. Runtime for kinematics does not vary much; in all cases, the forward and backward  account for about 21\% and 18\% of runtime for IKO and TO, respectively.
(4) \textbf{With large batch size, self-collision and collision take fewer percentages.}
Between the results for the Orin GPU running at 15W and MAXN power modes, the MAXN mode spends less of its runtime computing self-collision and collision cost, the variability of which leads to an 8\% to 10\% reduction in the runtime percentage of the depicted kernels for MAXN compared to 15W. The 2080 GPU, on the other hand, has a similar runtime breakdown across all environments, which most closely matches the breakdown of MAXN on the three greater-seed variants.

\subsection{Large Tensors Characterization}
\label{subsec:tensor-characterization}
Based on the characterization results, we analyze the most time-consuming CUDA kernels, including self-collision cost, collision cost, forward kinematics, and backward kinematics. Among these CUDA kernels, five large tensors cause significant memory latency. The output of forward kinematics: \textit{out\_spheres}, the input of backward kinematics: \textit{grad\_out\_spheres}, the input of self-collision cost: \textit{out\_vec}, the output of collision cost: \textit{closest\_pt} for IKO and \textit{closest\_pt\_swept} for TO.

\begin{figure}[t]
\centering
    \subfloat[][IKO.
    \label{fig:IKO_size}]{\includegraphics[width=0.48\columnwidth]{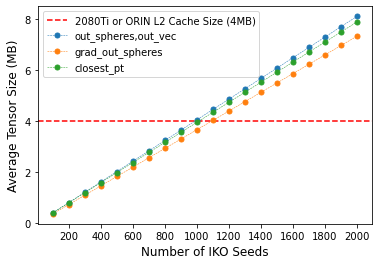}} 
    \subfloat[][TO.\label{fig:TO_size}]{\includegraphics[width=0.48\columnwidth]{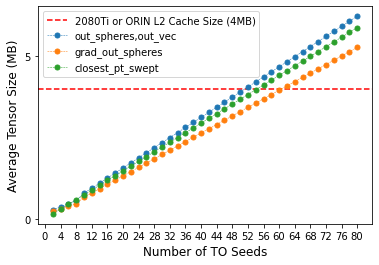}}
    \vspace{-0.08in}
    \caption{The average tensor sizes of IKO and TO iterations increase with the number of IKO and TO seeds, respectively.}
    \vspace{-0.22in}
    \label{fig:tensor-size}
\end{figure}

In Fig~\ref{fig:tensor-size}, we show the average tensor size with different IKO and TO seeds. We sweep the IKO seeds from 100 to 2000 and the TO seeds from 2 to 80. The tensor size grows linearly with the number of seeds. Both NVIDIA Orin and 2080 Ti have a 4MB L2 cache, and we label it as a dotted red line in both figures as a reference. All the tensors can grow larger than 4MB as the seeds increase. These large tensor data movements exceed the cache capacity and require off-chip DRAM memory, thus slowing down the CUDA kernels. If the tensor can fit into on-chip cache memory, the associated CUDA kernels reduce the expensive data transfers to the off-chip memory. 



Beyond tensor size, \textit{grad\_out\_spheres}, \textit{out\_vec}, \textit{closest\_pt}, and \textit{closest\_pt\_swept} have more than 99\% of sparsity. As a result, we adopt sparsity-aware computation by skipping zero computations. Since it is not the focus of this paper, we directly integrate the sparsity optimization into CuRobo.


\section{VaPr-CuRobo Overview}


\subsection{Search Space}
\label{subsec:search_space}
We start by identifying the size of the search space. 
Based on the characterization shown in Sec.~\ref{sec:characterization}, there are five large tensors that slow down the motion generation pipeline. Each tensor is in single-precision floating-point (FP) data format by default in the CuRobo pipeline. We aim to find optimized reduced-precision FP data types for these tensors without compromising the success rate of the original tasks. From a data movement perspective, we define the potential reduced-precision data formats, including FP4, FP5, FP6, FP8, FP10, FP16, and FP32. These data formats allow a 32-bit GPU register to store multiple reduced-precision FP scalars. A 32-bit GPU register can store one FP32, two FP16, three FP10, four FP8, five FP6, six FP5, or eight FP4, which reduces data transfers 2$\times$ to 8$\times$.

It is infeasible to perform an exhaustive search for the entire search space. Each FP data type can have different combinations of exponent and mantissa bits.
For instance, both E3M1 and E2M2 are FP5 data formats, where we define E3M1 as a FP data type of 3-bit exponent, 1-bit mantissa, and 1 sign bit, similarly for E2M2 and other variations. We consider 21 FP data types from FP4 to FP32,
which gives the search space $21^{5} = 4,084,101$. For each design point, it takes around 1 minute to evaluate 100 tasks with the CuRobo motion generation pipeline.

\subsection{VaPr Search Flow}
\begin{figure}[t!]
        \centering\includegraphics[width=0.9\columnwidth]{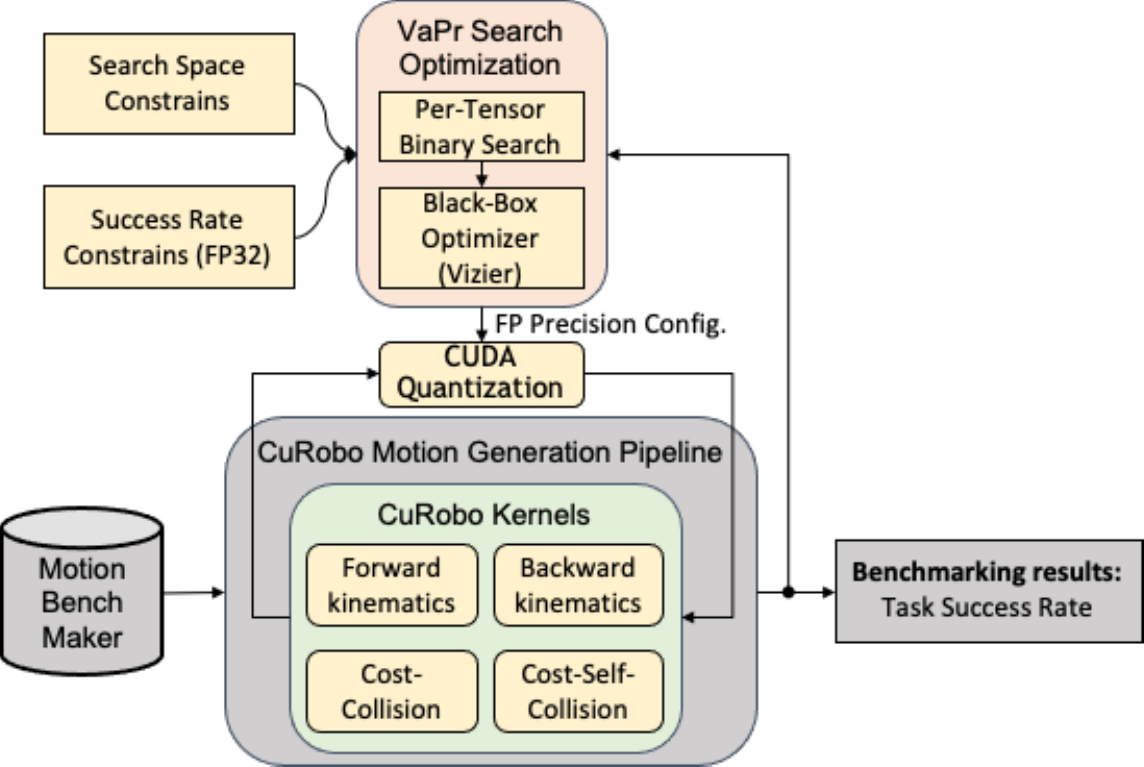}
        \caption{VaPr search technique overview.}
        \vspace{-0.25in}
        \label{fig:vapr-curobo}
\end{figure}

To find an optimized combination of reduced precision data formats out of the huge search space, we propose a VaPr search flow as shown in Fig.~\ref{fig:vapr-curobo}. The input to the VaPr flow includes search space constraints (described above) and success rate constraints.
The success rate constraints are to achieve the same success rate as CuRobo for the environments being considered. The VaPr search optimization block sends a FP precision configuration, including a FP data format for each tensor, to our custom CUDA quantization kernel. The CUDA quantization kernel will introduce numerical error for the large tensors at each kernel invocation by quantizing the tensors from FP32 to the specified data format and dequantizing them back to FP32. We evaluate the data format configuration by running the tasks in the provided environments and send the task success rate for each environment as a feedback signal to the search optimization.

\subsection{VaPr-CuRobo System Deployment}
VaPr-CuRobox can be easily deployed to a new environment. The VaPr search optimization is agnostic to the hardware and requires only a one-time search for a new environment. Moreover, for the 8 environments in MotionBenchMaker, our results show that 500 samples are enough to obtain a decent solution with VaPr search optimization, as shown in Sec.~\ref{subsec:variable-precision search}. The VaPr search technique supports parallelizing evaluation workload and thus could be finished within a few minutes in an industry setting with hundreds of machines.

\section{VaPr-CuRobo Details}
\label{sec:vapr}



\subsection{VaPr Search Optimization}
VaPr Search optimization is the most important component to find a set of feasible FP reduce-precision data formats for the tensors from the huge combinatorial search space. There are two phases during the search optimization: a per-tensor binary search and a combinatorial search across tensors. 

In the first phase, we reduce the combinatorial search space by reducing the possible FP data types for each tensor. We use binary search to find the minimum reduced-precision data format for each tensor while keeping the other tensors in the FP32 data format without success rate degradation. The per-tensor search results can thus eliminate all the FP data formats that are less than the corresponding result for each tensor. For instance, if the per-tensor search results show that FP16 is the minimum required data format for a tensor, any FP data formats that are less than 16 bits will not be considered, reducing the possible FP data formats from 21 combinations to 3 combinations (i.e., E5M10, E8M7, and E8M23) based on Sec.~\ref{subsec:search_space}. With fewer possibilities per tensor, the total search space across tensors decreases by 7.35$\times$, from 4,084,101 down to 555,660 for the \textit{table\_pick\_panda} and \textit{table\_under\_pick\_panda} environments, and by 7$\times$ for the remaining environments. 
The search space reduction is more for challenging environments or motion generation pipelines that require high precision.

In the second phase of the search optimization, we use Google Vizier~\cite{oss_vizier}, a black-box optimizer. We adopt NSGA-II~\cite{deb2002fast}, an evolutionary search algorithm, in the black-box optimizer. The objective of the black-box optimizer is to minimize the total number of bitwidths of the large tensors while satisfying the success rate constraints. The search space is the reduced search space filtered with per-tensor search results obtained in the first phase. At each optimization step, the black-box optimizer would sample from the reduced search space and provide reduced-precision FP data types for the tensors, including the FP data types with the specific number of exponent and mantissa bits. To evaluate the impact of reduced precision on the motion generation pipeline, our custom CUDA quantization kernel incurs numerical error based on the FP data types at each CUDA kernel invocation.

\subsection{FP Data Type Conversion}
To provide an out-of-box software solution, we leverage half-precision data type conversion supported by CUDA.
For the output tensors, we quantize every pair of FP32 values into a pair of half-precision FP16 (E5M10) values with CUDA intrinsic support, i.e., \textit{\_\_floats2half2\_rn}, and write out half of the original tensor. We dequantize every pair of FP16 (E5M10) values back to two FP32 values for the input tensor with \textit{\_\_half22float2}. Performance is expected to be similar with the use of the E8M7 16-bit floating point data type (a.k.a. bfloat16) with CUDA intrinsic support.
For other floating-point data types, we evaluate the performance benefits assuming hardware support for fast-type conversions. 

\section{Results}
\label{sec:result}
This section first shows the VaPr search results of the 8 environments from MotionBenchMaker~\cite{chamzas2021motionbenchmaker}. Based on the search results, we evaluate the performance gains for each CUDA kernel and the CuRobo motion generation pipeline. 

\subsection{VaPr Search}
\label{subsec:variable-precision search}
The VaPr search has two phases: per-tensor search and combinatorial search across tensors. Tab.~\ref{tab:per-tensor} shows the per-tensor search results for the large tensors across 8 environments. The table shows the minimum required bitwidth for each tensor while maintaining the rest of the motion generation pipeline at FP32. \textit{out\_spheres} requires from FP13 up to FP16 data format. \textit{grad\_out\_spheres}, \textit{out\_vec}, \textit{closest\_pt}, and \textit{closest\_pt\_swept} only need FP4 to FP5. The first observation is that FP32 is over-provisioned for these tensors and as low as FP4 is enough for the motion generation pipeline. The second observation is that the precision requirement varies for each tensor. \textit{out\_spheres} that represent the location of each limb requires more precision to avoid collision with the surrounding obstacles and itself. \textit{out\_vec}, \textit{closest\_pt}, and \textit{closest\_pt\_swept} are all regarding the cost for optimization. As long as the approximate magnitude and direction of the cost are correct. The optimization can continue to minimize the cost. Similarly, \textit{grad\_out\_spheres} is the input for backward propagation optimization. The bitwidth requirement doesn't vary much with the environments. More difficult environments require slightly higher precision, such as \textit{cage\_panda} and \textit{table\_under\_pick\_panda}.

\renewcommand{\arraystretch}{1.2}
\begin{table}
\caption{Results for 5 large tensors across 8 environments.}
\vspace{-5pt}
\resizebox{\columnwidth}{!}{
\begin{tabular}{|c|c|c|c|c|c|}
\hline
                          & out\_spheres & grad\_out\_spheres & out\_vec   & closest\_pt & closest\_pt\_swept \\ \hline \hline
\multicolumn{6}{|c|}{Per-tensor binary search} \\ \hline
                          
bookshelf\_small\_panda   & FP15 (E5M9)    & FP4 (E2M1)         & FP4 (E2M1) & FP4 (E2M1)  & FP4 (E2M1)         \\ \hline
bookshelf\_tall\_panda    & FP15 (E6M8)    & FP4 (E2M1)         & FP4 (E2M1) & FP4 (E2M1)  & FP4 (E2M1)         \\ \hline
bookshelf\_thin\_panda    & FP16 (E5M10)   & FP4 (E2M1)         & FP4 (E2M1) & FP4 (E2M1)  & FP4 (E2M1)         \\ \hline
box\_panda                & FP16 (E5M10)   & FP4 (E2M1)         & FP4 (E2M1) & FP4 (E2M1)  & FP4 (E2M1)         \\ \hline
box\_panda\_flipped       & FP16 (E5M10)   & FP4 (E2M1)         & FP4 (E2M1) & FP4 (E2M1)  & FP4 (E2M1)         \\ \hline
cage\_panda               & FP16 (E5M10)   & FP4 (E2M1)         & FP4 (E2M1) & FP4 (E2M1)  & FP4 (E2M1)         \\ \hline
table\_pick\_panda        & FP13 (E6M6)    & FP4 (E2M1)         & FP5 (E3M1) & FP4 (E2M1)  & FP4 (E2M1)         \\ \hline
table\_under\_pick\_panda & FP16 (E5M10)   & FP4 (E2M1)         & FP4 (E2M1) & FP4 (E2M1)  & FP5 (E3M1)         \\ \hline

\multicolumn{6}{|c|}{Combinatorial search}  \\ \hline
bookshelf\_small\_panda   & FP16 (E5M10) & FP8 (E4M3)         & FP4 (E2M1) & FP5 (E2M2)  & FP8 (E4M3)         \\ \hline
bookshelf\_tall\_panda    & FP16 (E5M10) & FP5 (E3M1)         & FP4 (E2M1) & FP4 (E2M1)  & FP5 (E3M1)         \\ \hline
bookshelf\_thin\_panda    & FP16 (E5M10) & FP6 (E2M3)         & FP5 (E3M1) & FP6 (E3M2)  & FP5 (E2M2)         \\ \hline
box\_panda                & FP16 (E5M10) & FP6 (E3M2)         & FP5 (E2M2) & FP6 (E3M2)  & FP5 (E3M1)         \\ \hline
box\_panda\_flipped       & FP16 (E5M10) & FP4 (E2M1)         & FP6 (E3M2) & FP5 (E2M2)  & FP5 (E3M1)         \\ \hline
cage\_panda               & FP16 (E5M10) & FP8 (E5M2)         & FP4 (E2M1) & FP4 (E2M1)  & FP6 (E2M3)         \\ \hline
table\_pick\_panda        & FP16 (E8M7)  & FP5 (E3M1)         & FP5 (E3M1) & FP5 (E2M2)  & FP6 (E3M2)         \\ \hline
table\_under\_pick\_panda & FP16 (E5M10) & FP6 (E3M2)         & FP5 (E2M2) & FP8 (E4M3)  & FP8 (E4M3)         \\ \hline
\end{tabular}
}
\label{tab:per-tensor}
\vspace{-0.2in}

\end{table}

Combinatorial search results are also shown in Tab.~\ref{tab:per-tensor}. We obtain the results by sampling 500 points with NSGA-II in VaPr search optimization. Compared to per-tensor search results, we see that more bits are required to maintain the success rate of the motion generation pipeline. It is likely that the numerical errors propagate among kernels, and the accumulated error is too large for the motion generation pipeline, especially in challenging tasks. The per-tensor search results can give a good starting point for the combinatorial search. However, we still need the second phase of the VaPr search optimization to search for a reduced-precision FP data types combination for the large tensors without compromising the motion generation pipeline. The final results show that we obtain 3.5$\times$ up to 4.4$\times$ tensor size reduction on average.

\subsection{Motion Generation Acceleration}
\label{subsec:motion_generation_acc}

\begin{figure*}[t!]
  \subfloat[Orin 15W.]{
	\begin{minipage}[b][][b]{
	   0.32\linewidth}
	   \includegraphics[width=\textwidth]{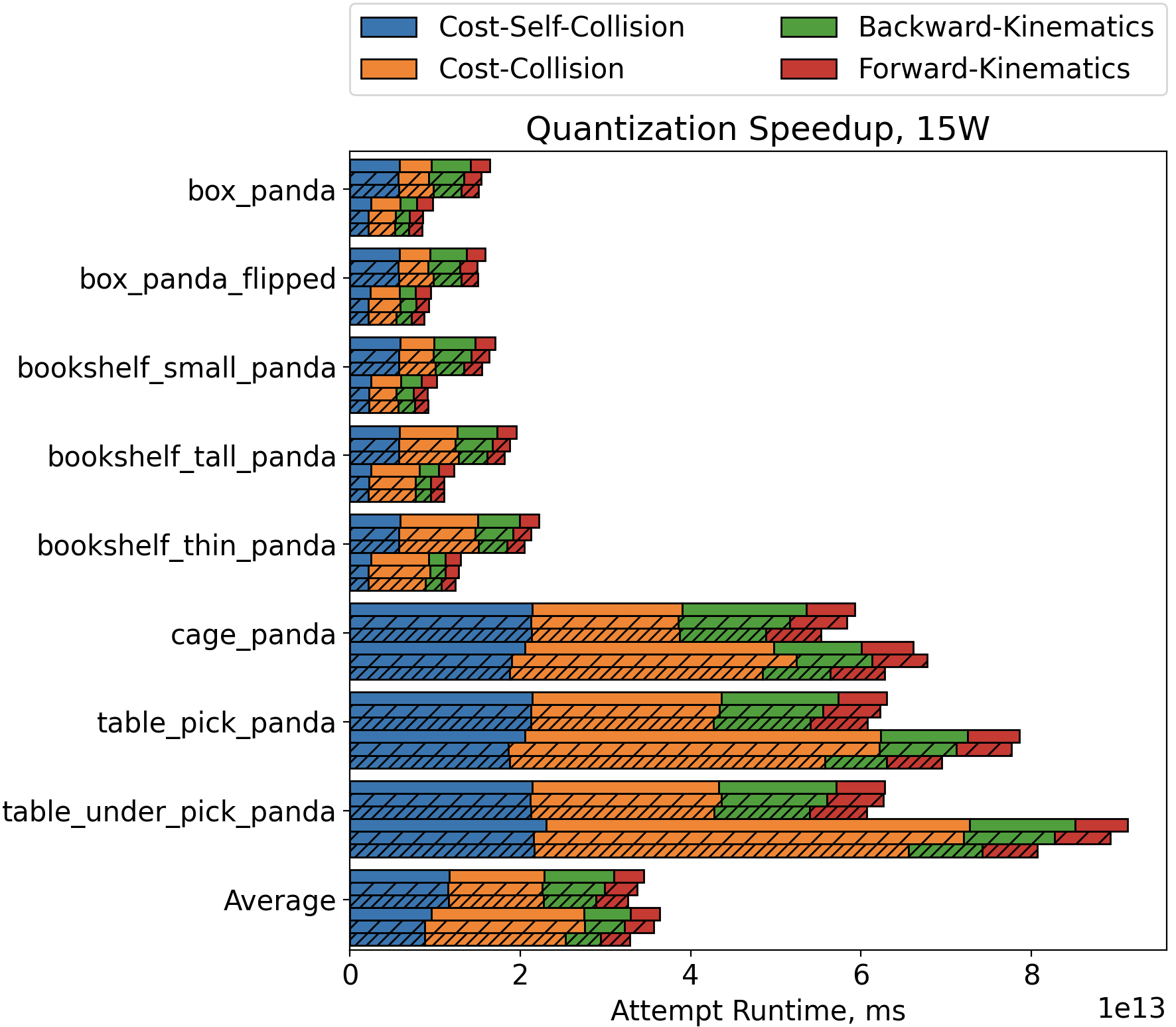}
	   \label{fig:15W_results}
	   \vspace{-0.2in}
	\end{minipage}}
	  \subfloat[Orin MAXN.]{
	\begin{minipage}[b][][b]{
	   0.32\linewidth}
	   \includegraphics[width=\textwidth]{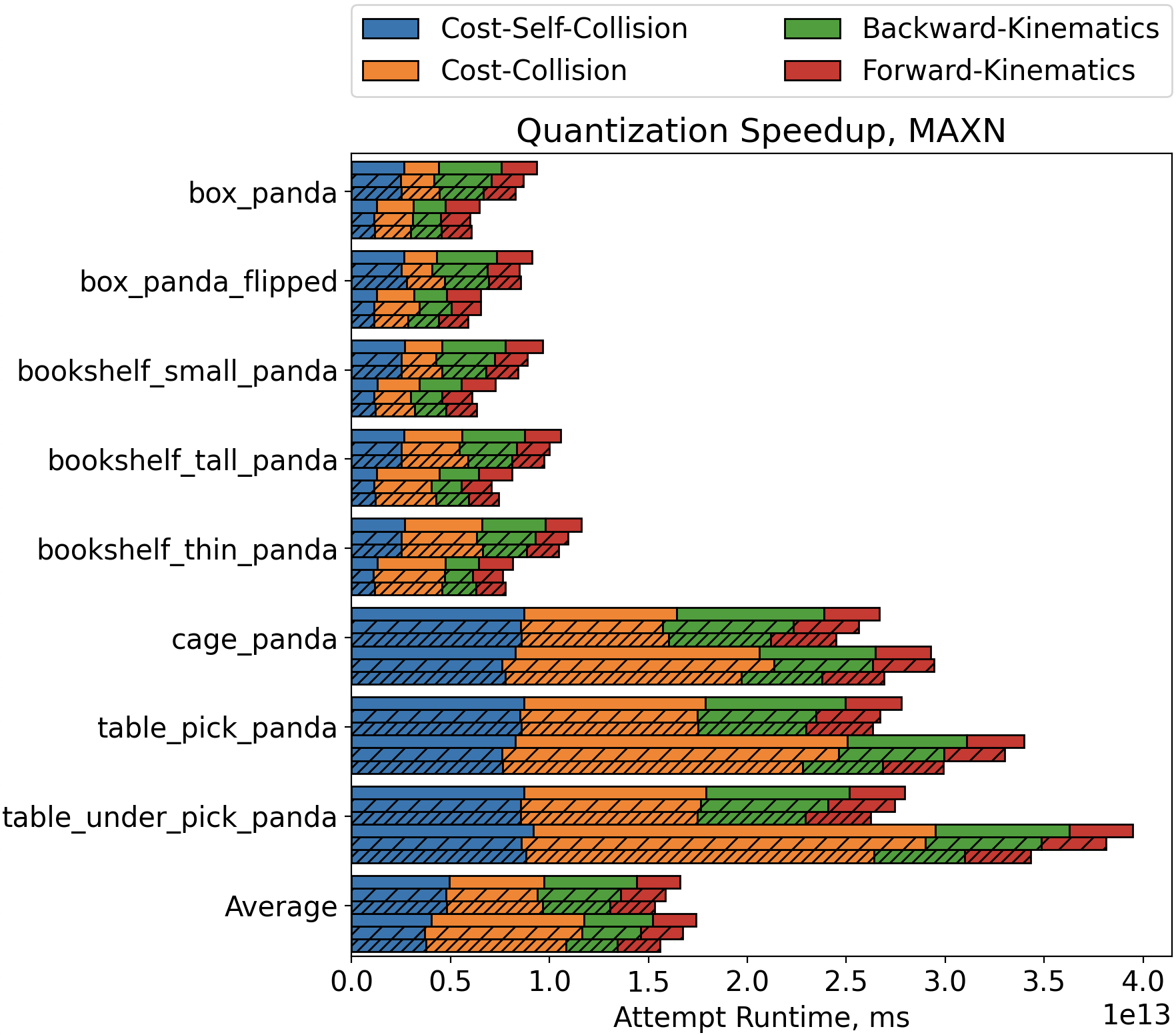}
	   \label{fig:MAXN_results}
	   \vspace{-0.2in}
	\end{minipage}}
		  \subfloat[2080 Ti.]{
	\begin{minipage}[b][][b]{
	   0.32\linewidth}
	   \includegraphics[width=\textwidth]{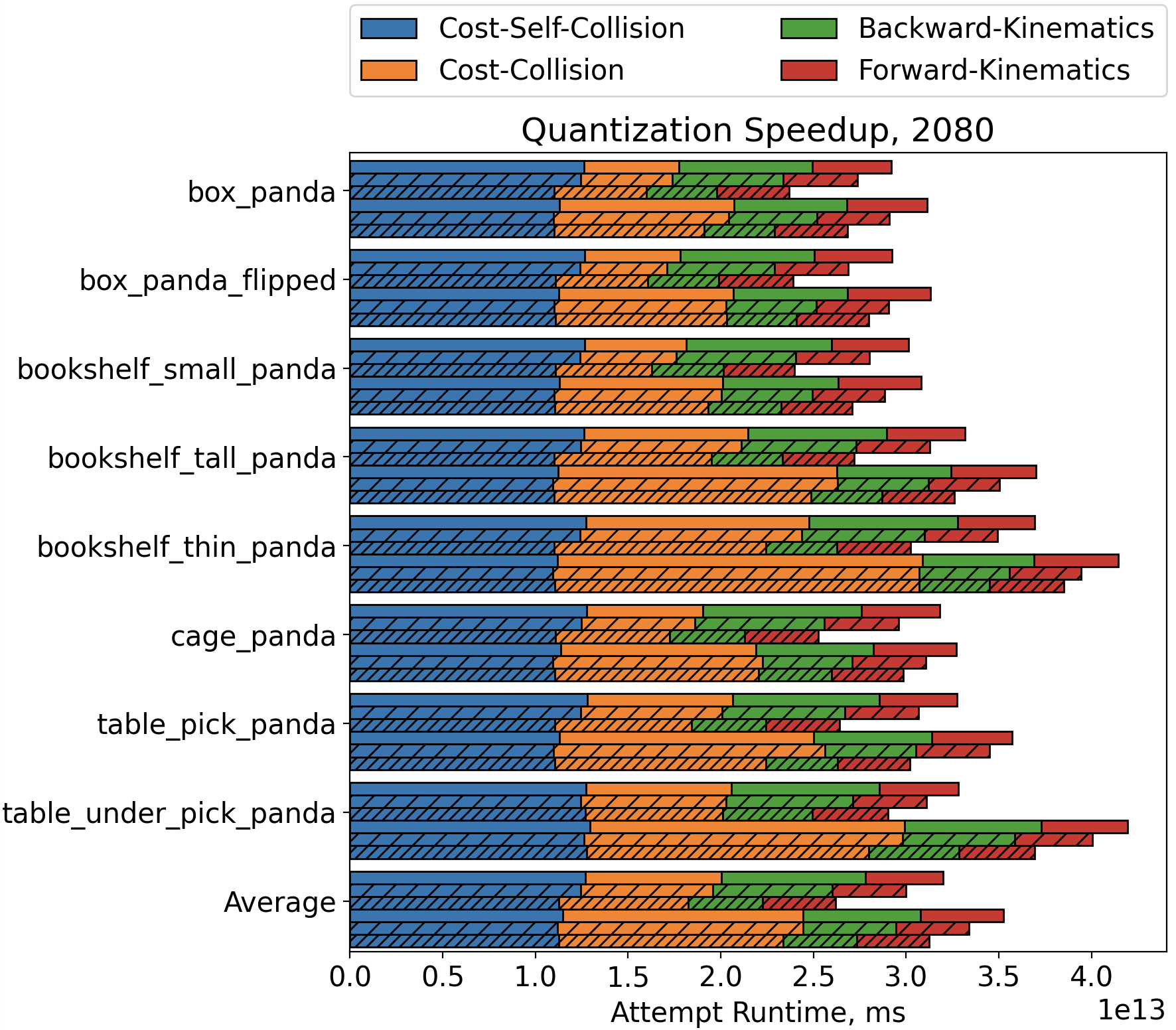}
	   \label{fig:2080_results}
	   \vspace{-0.2in}
	\end{minipage}}
	\vspace{-0.05in}
\caption{GPU runtime of one attempt of IKO 
at FP32, FP16, and VaPr precisions (upper three bars); TO at FP32, FP16, and VaPr precisions (lower three bars) on Orin and 2080 Ti GPU. Our proposed FP16 and VaPr techniques are in hatched bars.
}
\label{fig:results}
\vspace{-8pt}
\end{figure*}



\renewcommand{\arraystretch}{1.2}
\begin{table*}
\caption{Average and Maximum speedup with the proposed FP16 and VaPr.}
\vspace{-0.1in}
\resizebox{2\columnwidth}{!}{
\begin{tabular}{|c|c|c|c|c|c|c|}
\hline
           & \multicolumn{3}{|c|}{IKO}                   & \multicolumn{3}{|c|}{TO}                    \\ \hline \hline
(avg, max) & Orin 15 W        & Orin MAXN         & 2080 Ti            & Orin 15 W         & Orin MAXN         & 2080 Ti            \\ \hline
FP16       & \multicolumn{1}{c|}{(3.64\%, 6.54\%)} & \multicolumn{1}{c|}{(5.78\%, 8.98\%)}  & (6.85\%, 8.85\%)   & \multicolumn{1}{c|}{(5.23\%, 13.66\%)}  & \multicolumn{1}{c|}{(6.8\%, 19.14\%)}   & (5.73\%, 7.83\%)   \\ \hline
VaPr       & \multicolumn{1}{c|}{(6.84\%, 9.92\%)} & \multicolumn{1}{c|}{(9.31\%, 14.71\%)} & (22.32\%, 26.01\%) & \multicolumn{1}{c|}{(10.10\%, 14.04\%)} & \multicolumn{1}{c|}{(10.45\%, 15.10\%)} & (13.02\%, 18.16\%) \\ \hline
\end{tabular}
}
\label{tab:speedup}
\vspace{-0.05in}
\end{table*}



\renewcommand{\arraystretch}{1.2}
\begin{table*}
\caption{Median, 75 percentile, mean attempts for the baseline, half-precision with CUDA intrinsic, and Vapr precision.}
\vspace{-5pt}
\resizebox{2\columnwidth}{!}{
\begin{tabular}{|c|c|c|c|c|c|c|c|c|}
\hline
(median, 75\%, mean) Attempt & bookshelf\_small\_panda & bookshelf\_tall\_panda & bookshelf\_thin\_panda & box\_panda     & box\_panda\_flipped & cage\_panda    & table\_pick\_panda & table\_under\_pick\_panda \\ 
\hline
\hline
baseline (FP32)            & (1.0, 2, 2.35)          & (1.0, 2, 1.47)         & (1.0, 1, 1.68)         & (1.0, 2, 3.27) & (1.0, 2, 4.72)      & (1.0, 2, 2.56) & (1.0, 1, 2.23)     & (1.0, 1, 1.64)            \\ \hline
Half-precision               & (1.0, 2, 2.29)          & (1.0, 2, 2.36)         & (1.0, 1, 1.28)         & (1.0, 2, 4.69) & (1.0, 2, 3.38)      & (1.0, 2, 3.59) & (1.0, 1, 1.64)     & (1.0, 1, 1.48)            \\ \hline
Vapr                         & (1.0, 2, 2.90)          & (1.0, 1, 1.94)         & (1.0, 2, 1.77)         & (1.0, 1, 4.74) & (1.0, 3, 6.60)      & (1.0, 2, 3.22) & (1.0, 1, 1.35)     & (1.0, 3, 3.55)            \\ \hline
\end{tabular}
}
\label{tab:attempt}
\vspace{-0.2in}

\end{table*}
Figures~\ref{fig:results} shows per-attempt runtime for collision cost, self-collision cost, forward kinematics, and backward kinematics at FP32, FP16, and VaPr precisions for IK and TO across eight environments on Orin and 2080 Ti. The FP16 uses CUDA intrinsic function for type conversion between FP32 and FP16. The VaPr precision estimates the performance of fast type conversion assuming appropriate hardware support is available and quantizes the tensor based on the results shown in Tab.~\ref{tab:speedup}. There are 125 instances (kernel invocation) for a single attempt. 

On average, we achieve 4.43\%,	6.29\%,	and 6.29\% speedup with Orin 15W, Orin MAXN, and 2080 Ti with FP16, respectively, and 8.47\%,	9.88\%,	and 17.67\% speedup with Vapr on Orin 15W, MAXN, and 2080 Ti, respectively.

Tab.~\ref{tab:attempt} shows the number of attempts across the environments. The median attempt is one across all settings. The 75 percentile attempts are also similar for the three settings. While the mean attempt is slightly more for VaPr, it can be decreased by adjusting the hyperparameters in CuRobo or by relying on better seeding algorithms (e.g., DNN-based trajectory predictors) to generate a better initialization.

\section{Related Work}
\label{related}

Model predictive control methods~\cite{farshidian2017real,bhardwaj2022storm
} are shown to reach real-time solutions. However, they are only optimized locally, and the generated trajectories can be sub-optimal. Some methods use DNNs for motion planning~\cite{qureshi2019motion
}, reaching fast but not yet real-time solutions (around 80ms per trajectory). 
%
%
GRiD~\cite{plancher2021grid} is a GPU-accelerated library that generates CUDA kernels for rigid body dynamics. 
VaPr can be extended beyond CuRobo, e.g., to rigid-body dynamics.

While mixed precision quantization to accelerate motion planning has not been studied, our high-level approach has similarities to mixed precision per-layer quantization in DNNs~\cite{
sakrIcassp2018
}. 
\section{Conclusion}
\label{sec:conclusion}
We propose a reduced precision search technique to find the minimum precision for motion generation pipelines at the per-tensor level. We evaluated our approach on a difficult set of motion
generation problems for manipulators~\cite{chamzas2021motionbenchmaker} and show an average of 6.3\% and 6.3\%
speedup with CUDA intrinsics support, and 9.9\% and 17.7\% speedup with the proposed fast type conversion hardware over the SOTA GPU accelerated motion generation method on Jetson Orin and RTX 2080 Ti GPU, respectively.
\bibliographystyle{IEEEtranS}
\bibliography{refs}
\section*{Appendix}

\section{Experimental Setup for VaPr Performance Estimation}

We describe the experimental setup for VaPr performance estimation introduced in Sec.~IV-B of the paper.
Based on the paper's precision search results in Tab.~II, the highest precision requirements across the environments are FP16, FP8, FP6, FP8, and FP8 for the five tensors under consideration. For FP16, we leverage CUDA intrinsic type conversion between FP32 and FP16. Since there is no CUDA intrinsic type conversion support below FP16, we estimate memory transfer reduction of quantization by transferring fewer FP32 values. For the other four tensors, we conservatively move one float out of every three floats, which has a similar effect of quantization from FP32 to FP10 regarding memory movements. We add two addition operations for every three floats to mimic the time of a fast-type conversion instruction.


\section{Detailed Experimental Results}

In addition to Tab.~III in the paper, we present the speedup for each environment with the two options -- FP16 CUDA intrinsics support in Tab.~\ref{tab:fp16} and VaPr in Tab.~\ref{tab:vapr}. The speedup is defined as $ ( (\textit{FP32 baseline} \quad / \quad \textit{FP16 or vapr}) - 1) * 100\% $.
\renewcommand{\arraystretch}{1.2}

\begin{table}[h]
\centering
\caption{Speedup with the use of FP16 data type compared to the FP32 baseline.}
\resizebox{\linewidth}{!}{
\begin{tabular}{|c|ccc|ccc|}
\hline
\multirow{2}{*}{FP16}       & \multicolumn{3}{c|}{IKO}                                                                          & \multicolumn{3}{c|}{TO}                                                                            \\ \cline{2-7} 
                            & \multicolumn{1}{c|}{Orin 15 W}        & \multicolumn{1}{c|}{Orin MAXN}        & 2080 Ti          & \multicolumn{1}{c|}{Orin 15 W}         & \multicolumn{1}{c|}{Orin MAXN}         & 2080 Ti          \\ \hline \hline
bookshelf\_small\_panda   & \multicolumn{1}{c|}{4.30\%}    & \multicolumn{1}{c|}{8.98\%}    & 7.55\%  & \multicolumn{1}{c|}{12.62\%}   & \multicolumn{1}{c|}{19.14\%}   & 6.80\%  \\ \hline
bookshelf\_tall\_panda    & \multicolumn{1}{c|}{4.07\%}    & \multicolumn{1}{c|}{5.83\%}    & 6.07\%  & \multicolumn{1}{c|}{10.92\%}   & \multicolumn{1}{c|}{15.16\%}   & 5.58\%  \\ \hline
bookshelf\_thin\_panda    & \multicolumn{1}{c|}{4.69\%}    & \multicolumn{1}{c|}{6.33\%}    & 5.71\%  & \multicolumn{1}{c|}{1.70\%}    & \multicolumn{1}{c|}{6.73\%}    & 5.05\%  \\ \hline
box\_panda                & \multicolumn{1}{c|}{6.48\%}    & \multicolumn{1}{c|}{7.70\%}    & 6.65\%  & \multicolumn{1}{c|}{13.66\%}   & \multicolumn{1}{c|}{7.49\%}    & 7.01\%  \\ \hline
box\_panda\_flipped       & \multicolumn{1}{c|}{6.54\%}    & \multicolumn{1}{c|}{7.34\%}    & 8.85\%  & \multicolumn{1}{c|}{1.84\%}    & \multicolumn{1}{c|}{-0.13\%}   & 7.83\%  \\ \hline
cage\_panda               & \multicolumn{1}{c|}{1.54\%}    & \multicolumn{1}{c|}{4.13\%}    & 7.57\%  & \multicolumn{1}{c|}{-2.38\%}   & \multicolumn{1}{c|}{-0.50\%}   & 5.35\%  \\ \hline
table\_pick\_panda        & \multicolumn{1}{c|}{1.32\%}    & \multicolumn{1}{c|}{4.10\%}    & 6.82\%  & \multicolumn{1}{c|}{1.19\%}    & \multicolumn{1}{c|}{2.99\%}    & 3.50\%  \\ \hline
table\_under\_pick\_panda & \multicolumn{1}{c|}{0.16\%}    & \multicolumn{1}{c|}{1.81\%}    & 5.56\%  & \multicolumn{1}{c|}{2.31\%}    & \multicolumn{1}{c|}{3.49\%}    & 4.73\%  \\ \hline 
All environments avg      & \multicolumn{1}{c|}{3.64\%}    & \multicolumn{1}{c|}{5.78\%}    & 6.85\%  & \multicolumn{1}{c|}{5.23\%}    & \multicolumn{1}{c|}{6.80\%}    & 5.73\%  \\ \hline
All environments max      & \multicolumn{1}{c|}{6.54\%}    & \multicolumn{1}{c|}{8.98\%}    & 8.85\%  & \multicolumn{1}{c|}{13.66\%}   & \multicolumn{1}{c|}{19.14\%}   & 7.83\%  \\ \hline
\end{tabular}
}

\label{tab:fp16}
\vspace{-20pt}
\end{table}
\renewcommand{\arraystretch}{1.2}

\begin{table}[h]
\centering
\caption{Speedup with VaPr compared to the FP32 baseline.}
\resizebox{\linewidth}{!}{
\begin{tabular}{|c|ccc|ccc|}
\hline
\multirow{2}{*}{VaPr}       & \multicolumn{3}{c|}{IKO}                                                                             & \multicolumn{3}{c|}{TO}                                                                              \\ \cline{2-7} 
                            & \multicolumn{1}{c|}{Orin 15 W}        & \multicolumn{1}{c|}{Orin MAXN}         & 2080 Ti            & \multicolumn{1}{c|}{Orin 15 W}         & \multicolumn{1}{c|}{Orin MAXN}         & 2080 Ti            \\ \hline \hline
bookshelf\_small\_panda   & \multicolumn{1}{c|}{9.92\%}    & \multicolumn{1}{c|}{14.71\%}   & 25.68\% & \multicolumn{1}{c|}{11.05\%}   & \multicolumn{1}{c|}{15.10\%}   & 13.79\% \\ \hline
bookshelf\_tall\_panda    & \multicolumn{1}{c|}{7.79\%}    & \multicolumn{1}{c|}{8.61\%}    & 21.99\% & \multicolumn{1}{c|}{10.70\%}   & \multicolumn{1}{c|}{9.40\%}    & 13.43\% \\ \hline
bookshelf\_thin\_panda    & \multicolumn{1}{c|}{8.63\%}    & \multicolumn{1}{c|}{10.85\%}   & 22.16\% & \multicolumn{1}{c|}{5.30\%}    & \multicolumn{1}{c|}{4.78\%}    & 7.66\%  \\ \hline
box\_panda                & \multicolumn{1}{c|}{8.49\%}    & \multicolumn{1}{c|}{12.70\%}   & 23.14\% & \multicolumn{1}{c|}{14.04\%}   & \multicolumn{1}{c|}{6.41\%}    & 16.08\% \\ \hline
box\_panda\_flipped       & \multicolumn{1}{c|}{5.56\%}    & \multicolumn{1}{c|}{6.51\%}    & 22.27\% & \multicolumn{1}{c|}{8.06\%}    & \multicolumn{1}{c|}{10.47\%}   & 11.95\% \\ \hline
cage\_panda               & \multicolumn{1}{c|}{7.08\%}    & \multicolumn{1}{c|}{9.03\%}    & 26.01\% & \multicolumn{1}{c|}{5.35\%}    & \multicolumn{1}{c|}{8.82\%}    & 9.53\%  \\ \hline
table\_pick\_panda        & \multicolumn{1}{c|}{3.75\%}    & \multicolumn{1}{c|}{5.57\%}    & 24.12\% & \multicolumn{1}{c|}{13.18\%}   & \multicolumn{1}{c|}{13.62\%}   & 18.16\% \\ \hline
table\_under\_pick\_panda & \multicolumn{1}{c|}{3.46\%}    & \multicolumn{1}{c|}{6.48\%}    & 13.18\% & \multicolumn{1}{c|}{13.09\%}   & \multicolumn{1}{c|}{14.97\%}   & 13.55\% \\ \hline
All environments avg      & \multicolumn{1}{c|}{6.84\%}    & \multicolumn{1}{c|}{9.31\%}    & 22.32\% & \multicolumn{1}{c|}{10.10\%}   & \multicolumn{1}{c|}{10.45\%}   & 13.02\% \\ \hline
All environments max      & \multicolumn{1}{c|}{9.92\%}    & \multicolumn{1}{c|}{14.71\%}   & 26.01\% & \multicolumn{1}{c|}{14.04\%}   & \multicolumn{1}{c|}{15.10\%}   & 18.16\% \\ \hline
\end{tabular}
}

\label{tab:vapr}
\end{table}

Similarly, we present the speedup for the targeted kernels with FP16 in Tab.~\ref{tab:kernel_fp16} and VaPr in Tab.~\ref{tab:kernel_vapr}. 
\renewcommand{\arraystretch}{1.2}

\begin{table}[h]
\centering
\caption{Per-kernel speedup with the use of FP16 data type compared to the FP32 baseline.}
\resizebox{\linewidth}{!}{
\begin{tabular}{|c|ccc|ccc|}
\hline
\multirow{2}{*}{FP16} & \multicolumn{3}{c|}{IKO}                                                  & \multicolumn{3}{c|}{TO}                                                   \\ \cline{2-7} 
                      & \multicolumn{1}{c|}{Orin 15 W} & \multicolumn{1}{c|}{Orin MAXN} & 2080 Ti & \multicolumn{1}{c|}{Orin 15 W} & \multicolumn{1}{c|}{Orin MAXN} & 2080 Ti \\ \hline
Forward kinematics    & \multicolumn{1}{c|}{0.77\%}    & \multicolumn{1}{c|}{1.59\%}    & 5.76\%  & \multicolumn{1}{c|}{7.61\%}    & \multicolumn{1}{c|}{6.23\%}    & 13.47\% \\ \hline
Backward kinematics   & \multicolumn{1}{c|}{11.06\%}   & \multicolumn{1}{c|}{10.81\%}   & 20.96\% & \multicolumn{1}{c|}{16.71\%}   & \multicolumn{1}{c|}{18.79\%}   & 27.04\% \\ \hline
Cost-Self-Collision   & \multicolumn{1}{c|}{1.70\%}    & \multicolumn{1}{c|}{4.67\%}    & 2.23\%  & \multicolumn{1}{c|}{10.15\%}   & \multicolumn{1}{c|}{11.81\%}   & 2.73\%  \\ \hline
Cost-Collision        & \multicolumn{1}{c|}{1.39\%}    & \multicolumn{1}{c|}{3.93\%}    & 3.17\%  & \multicolumn{1}{c|}{-1.83\%}   & \multicolumn{1}{c|}{-2.06\%}   & -2.30\% \\ \hline
\end{tabular}
}

\label{tab:kernel_fp16}
\end{table}
\renewcommand{\arraystretch}{1.2}

\begin{table}[h]
\centering
\caption{Per-kernel speedup with VaPr compared to the FP32 baseline.}
\resizebox{\linewidth}{!}{
\begin{tabular}{|c|ccc|ccc|}
\hline
\multirow{2}{*}{FP16} & \multicolumn{3}{c|}{IKO}                                                  & \multicolumn{3}{c|}{TO}                                                   \\ \cline{2-7} 
                      & \multicolumn{1}{c|}{Orin 15 W} & \multicolumn{1}{c|}{Orin MAXN} & 2080 Ti & \multicolumn{1}{c|}{Orin 15 W} & \multicolumn{1}{c|}{Orin MAXN} & 2080 Ti \\ \hline
Forward kinematics    & \multicolumn{1}{c|}{1.17\%}    & \multicolumn{1}{c|}{1.29\%}    & 6.50\%  & \multicolumn{1}{c|}{7.33\%}    & \multicolumn{1}{c|}{5.44\%}    & 13.95\% \\ \hline
Backward kinematics   & \multicolumn{1}{c|}{36.13\%}   & \multicolumn{1}{c|}{39.60\%}   & 95.38\% & \multicolumn{1}{c|}{24.64\%}   & \multicolumn{1}{c|}{26.07\%}   & 60.14\% \\ \hline
Cost-Self-Collision   & \multicolumn{1}{c|}{1.21\%}    & \multicolumn{1}{c|}{3.10\%}    & 13.15\% & \multicolumn{1}{c|}{10.48\%}   & \multicolumn{1}{c|}{8.24\%}    & 2.04\%  \\ \hline
Cost-Collision        & \multicolumn{1}{c|}{-2.78\%}   & \multicolumn{1}{c|}{-4.96\%}   & 4.19\%  & \multicolumn{1}{c|}{5.75\%}    & \multicolumn{1}{c|}{6.34\%}    & 7.60\%  \\ \hline
\end{tabular}
}

\label{tab:kernel_vapr}
\end{table}

\end{document}